\documentclass[review]{elsarticle}

\usepackage{algorithm} 
\usepackage{etoolbox,lipsum}
\usepackage{amsmath}
\usepackage{algorithmic}
\usepackage{xcolor, color}
\usepackage{array, makecell, multirow}
\usepackage{amssymb}

\usepackage{multirow}
\usepackage{tabularx}

\usepackage{xcolor}

\usepackage{colortbl}

\journal{Pattern Recognition}

\begin{document}

\begin{frontmatter}



\title{Audio-Driven Talking Face Generation with Diverse yet Realistic Facial Animations}


\author[address1]{Rongliang Wu}
\author[address1]{Yingchen Yu}
\author[address2]{Fangneng Zhan}
\author[address1]{Jiahui Zhang}
\author[address3]{Xiaoqin Zhang}
\author[address1]{Shijian Lu\corref{mycorrespondingauthor}}
\cortext[mycorrespondingauthor]{Corresponding author \\
                                Email Address: \{ronglian001, yingchen001, jiahui003\}@e.ntu.edu.sg, 
                                fzhan@mpi-inf.mpg.de, 
                                zhangxiaoqinnan@gmail.com,  
                                shijian.lu@ntu.edu.sg}

\address[address1]{School of Computer Science and Engineering, Nanyang Technological University, Singapore}
\address[address2]{Max Planck Institute for Informatics, Germany}
\address[address3]{College of Computer Science and Artifcial Intelligence, Wenzhou University, China}

\begin{abstract}
Audio-driven talking face generation, which aims to synthesize talking faces with realistic facial animations (including accurate lip movements, vivid facial expression details and natural head poses) corresponding to the audio, has achieved rapid progress in recent years.
However, most existing work focuses on generating lip movements only without handling the closely correlated facial expressions, which degrades the realism of the generated faces greatly. 
This paper presents DIRFA, a novel method that can generate talking faces with diverse yet realistic facial animations from the same driving audio. 
To accommodate fair variation of plausible facial animations for the same audio, we design a transformer-based probabilistic mapping network that can model the variational facial animation distribution conditioned upon the input audio and autoregressively convert the audio signals into a facial animation sequence.
In addition, we introduce a temporally-biased mask into the mapping network, which allows to model the temporal dependency of facial animations and produce temporally smooth facial animation sequence. 
With the generated facial animation sequence and a source image, photo-realistic talking faces can be synthesized with a generic generation network.
Extensive experiments show that DIRFA can generate talking faces with realistic facial animations effectively.
\end{abstract}


\end{frontmatter}


\section{Introduction}
\label{sec:intro}

Audio-driven talking face generation aims to synthesize talking faces with realistic facial animations that correspond to the input audio signals, which has attracted increasing interest from both academia and industry due to its wide applications in digital human, visual dubbing, virtual reality, etc. 
Facial animations are usually closely entangled with the corresponding speech, which contain strong non-verbal information that helps the audience understand the speech contents~\cite{cassell1999speech}. 
Generating naturally-looking facial animations is therefore one key factor in realistic talking face generation, which remains a very open research challenge. Specifically, modelling facial animations from audio signals is not a rigid one-to-one mapping problem, which requires to accommodate fair variations of plausible facial animations given the same audio. 

Audio-driven talking face generation has been explored extensively with the prevalence of deep generative networks~\cite{goodfellow2014generative} in recent years. 
One typical approach is static talking face generation~\cite{fan2015photo, chen2018lip, prajwal2020lip} which edits lip movements only without considering other facial animations (i.e., head poses and facial expressions). Another approach focuses on dynamic talking face generation~\cite{chen2020talking, yi2020audio, zhou2021pose}, which includes head movements for modelling full-face animations but the generated faces are still expressionless. Hence, most existing work focuses on generating lip-synced talking faces with respect to the driving audio (with or without head movements) but neglects facial expressions which are crucial to realistic talking face generation.

At the other end, generating realistic audio-driven facial expressions is a non-trivial task. 
This is largely due to the fact that facial expressions do not have very strong correlations with the corresponding audio. Specifically, there often exist fair variations of plausible facial expressions for the same audio signal, and the variations might further expand while extending to the temporal dimension with a sequence of audio signals as input. To model such uncertainty, the desired network should be capable of accommodating fair variations of plausible facial expressions for the same input audio. Deterministic prediction of facial expressions from the input audio will drive the regression to the mean facial expression which often leads to expressionless talking face videos~\cite{chen2020talking, yi2020audio, zhou2021pose}.

This paper presents \textbf{DIRFA}, an innovative audio-driven talking face generation method that can generate talking faces with \textbf{DI}verse yet \textbf{R}ealistic \textbf{F}acial \textbf{A}nimations (including \textit{accurate} lip movements, \textit{vivid} facial expression details and \textit{natural} head poses) from the same driving audio. 
Specifically, we design a mapping network to model the uncertain relations between audio and visual signals. 
The design is inspired by the transformer architecture~\cite{vaswani2017attention}, which allows the mapping network to model the variational facial animation distribution conditioned upon the input audio and convert the audio into facial animation sequence in an autoregerssive manner. 
In addition, we introduce a temporally-biased mask into the mapping network, which allows to model the temporal dependency of facial animations and produce temporally coherent animation sequences effectively by assigning higher attention weights to closer facial frames while generating new animations.
With the generated facial animation sequence and a source facial image, talking face videos with realistic facial animations can be synthesized with a generic generation network.

The contributions of this work can be summarized in three aspects.
First, we propose DIRFA, a novel audio-driven talking face generation method that can generate talking face videos with diverse yet realistic facial animations from the same input audio. 
Second, we design a probabilistic mapping network that can model the uncertain relations between audio and visual signals effectively. In addition, we introduce a temporally-biased mask into the mapping network, which enables it to generate temporally coherent facial animations.
Third, extensive experiments show that DIRFA can generate realistic talking face videos with naturally-looking facial animations.

\section{Related Work}

\subsection{Talking Face Generation}

\textbf{Audio-driven Talking Face Generation: } 
Audio-driven talking face generation has been studied in both computer graphics and computer vision communities for years.  
Early studies focus on subject-specific talking face synthesis. For example, Suwajanakorn et al.~\cite{suwajanakorn2017synthesizing} generate mouth movement from the driving audio and composite it with video frames of the same speaker to synthesize talking faces. This approach requires a large amount of video footage of one speaker to train a speaker-specific model that cannot generalize to new persons. 
Recently, several studies~\cite{song2018talking, chen2019hierarchical, zhou2019talking, zhou2020makelttalk, prajwal2020lip, vougioukas2020realistic, liu2020synthesizing} exploit deep generative networks for subject-independent talking face generation. For example, Chen et al.~\cite{chen2019hierarchical} present a hierarchical structure that first predicts facial landmarks from the audio and then generates talking faces conditioned on the predicted landmarks. However, all aforementioned methods generate head-fixed talking faces which degrades the  realism of the synthesized videos greatly. To improve the perceptual realism, some recent work~\cite{chen2020talking, yi2020audio, zhou2021pose, wang2021audio2head, zhang2021facial} considers head poses while generating talking face videos.

Although modelling head movements from speech improves the realism of talking faces, most existing methods share a common constraint that they usually synthesize expressionless talking faces as they neglect to model facial expressions from the audio. This directly degrades the realism of the generated talking faces as human facial expressions usually change with the speech. 
Leveraging the dedicatedly collected data, some recent studies~\cite{ji2021audio,wang2020mead} attempt to control the facial expressions of the generated talking faces of a specific person, but their models cannot generalize to new persons. Liang et al.~\cite{liang2022expressive} propose to use an additional emotional video for guiding expression synthesis.
Our proposed DIRFA instead can synthesize talking faces with diverse yet realistic facial expressions from audio only and the trained model is generalizable to unseen persons.

\textbf{Video-driven Talking Face Generation: } 
Video-driven talking face generation aims to transfer facial animations from a reference actor to a target person. For example, Zakharov et al.~\cite{zakharov2019few} propose a few-shot talking head model that generates talking faces conditioned on the facial landmarks.
Ren et al.~\cite{ren2021pirenderer} leverage 3D face models to transfer facial animations from the reference videos to another person.
Wiles et al.~\cite{wiles2018x2face} introduce a warping-based talking face generation network. 
Although these methods can generate photo-realistic talking face videos, they require videos with desired facial animations to guide the synthesis.

\subsection{Modelling Facial Animations from Audio}

One fundamental challenge in audio-driven talking face generation is how to accurately convert audio contents into visual signals. Existing methods tackle this challenge by either implicitly modelling facial animations from audio with latent features~\cite{song2018talking, zhou2019talking} or explicitly converting audio into intermediate visual representations such as 2D facial landmarks~\cite{chen2019hierarchical, zhou2020makelttalk} and 3D face coefficients~\cite{zhang2021facial, ren2021pirenderer}. Though great progress has been witnessed in generating accurate lip sync, most existing work models audio-to-visual mapping in a deterministic manner and ignores the uncertainty between audio and facial expressions. This leads to regress-to-mean problem~\cite{wang2021audio2head} where the synthesized faces have little variation in facial expressions.

Differently, our proposed mapping network is trained to model the variational facial animation distribution conditioned upon the input audio, which allows to convert the audio signals into facial animation sequences with diverse yet realistic facial animations. Specifically, we map audio signals to facial animations by exploiting continuous Action Units (AUs)~\cite{ekman2002facial} for jointly modelling mouth movements and facial expressions, and a rigid 6 degrees of freedom (DoF) movement vector (pitch, yaw, roll and 3D translation) for modelling head pose.

\subsection{Transformers in Vision}

Transformer~\cite{vaswani2017attention} emerges as a powerful tool for modelling long-range contextual information, which has been widely studied in natural language processing. 
The core of transformer is the attention mechanism that allows for interaction between sequences regardless of the relative positions. 
A basic transformer building block consists of a multi-head attention-layer (Attn) followed by a feed forward layer (FF), which embeds input sequence $X$ into an internal representation $C$, which is often referred to as the context vector~\cite{li2021learn}.
Specifically, $C$ is computed using the query $Q$, the key $K$ and the value $V$ via:
\begin{equation}
\small
    \begin{aligned}
        C = FF(Attn(Q, K, V, M)) & = FF(softmax(\frac{QK^T + M}{\sqrt{N_k}})V), \\
        Q = XW^Q, K & = XW^K, V=XW^V, 
    \end{aligned}
    \label{transformer_basic}
\end{equation}
where $N_k$ is the number of channels, $Ws$ are trainable weights and $M$ is a mask that enables causal attention, where each token can only attend to past inputs~\cite{radford2018improving}.

Dosovitskiy et al.~\cite{dosovitskiy2020image} make the first attempt to apply transformer to image classification, which shows the potential of transformer in solving computer vision problems.
From then on, transformer-based models have been exploited for different tasks, including object detection~\cite{carion2020end}, image generation~\cite{esser2021taming}), video understanding~\cite{girdhar2019video}, etc.
Some recent work explores applying transformer to model sequential data and produces impressive results on motion synthesis~\cite{aksan2021spatio} and dance generation~\cite{li2021learn, siyao2022bailando}. 
Nevertheless, most existing methods either focus on motion generation without considering audio information, or cannot generate diverse motion sequences given audio only.
In this paper, we employ transformer to convert audio signals into diverse yet realistic facial animation sequences for talking face generation.

\begin{figure*}[!t]
\begin{center}
\includegraphics[width=1.\linewidth]{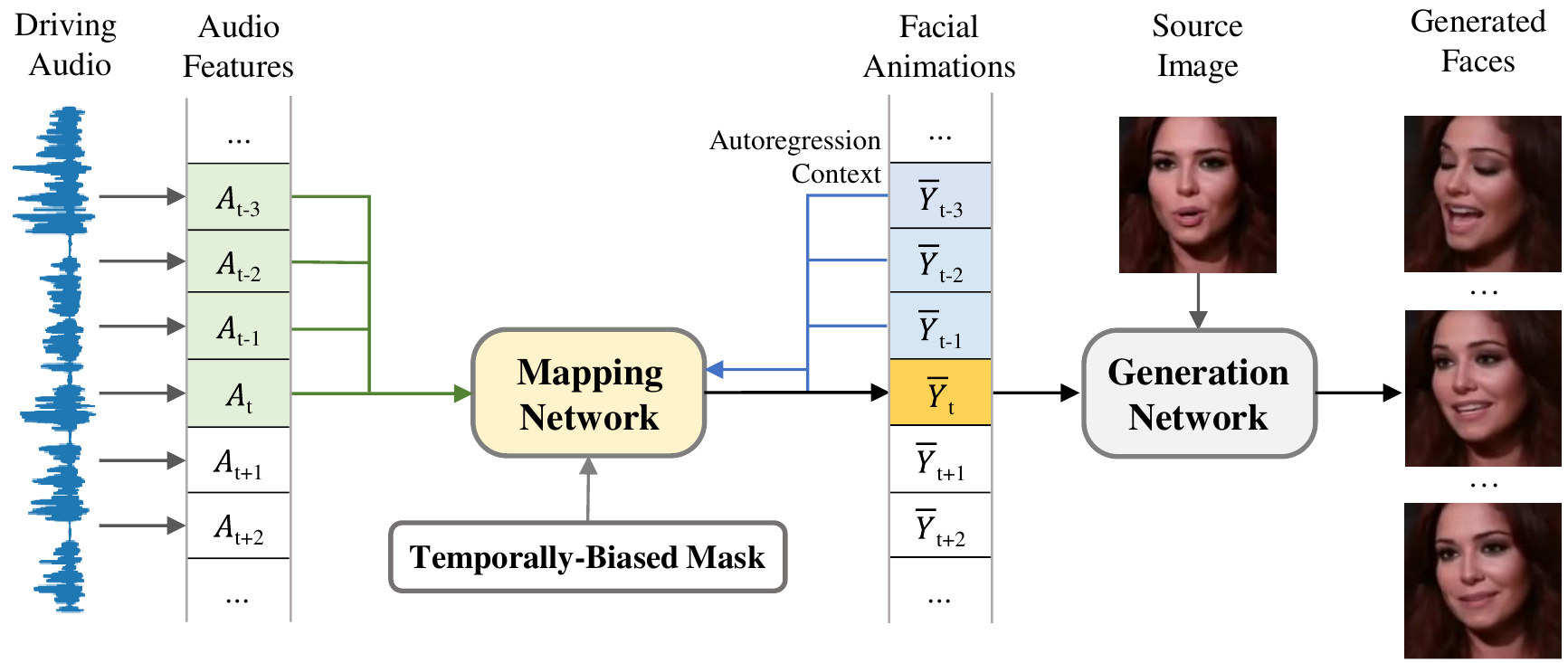}
\end{center}
\caption{
    The framework of DIRFA: Given the driving audio, we first convert it into audio features that are temporally aligned with the video frames. The audio features are then fed to a mapping network that generates plausible facial animations in an autoregressive manner. A temporally-biased mask is introduced to the mapping network to enable it to better model the temporal dependency of facial animations and produce temporally coherent facial animation sequence. Finally, the generated facial animations and a source image are forwarded to a generation network to synthesize talking faces with \textit{accurate} lip movements, \textit{vivid} facial expressions and \textit{natural} head poses.
}
\label{fig:pipeline}
\end{figure*}

\section{Method}

\subsection{Overview}

Fig.~\ref{fig:pipeline} shows the framework of our proposed DIRFA. 
As audio usually has strong correlations with lip shapes and relatively weak correlations with facial expressions and head poses, our goal is to generate talking faces with \textit{accurate} lip movements, \textit{vivid} facial expression details and \textit{natural} head poses corresponding to the driving audio.
To achieve this goal, we design a probabilistic mapping network that exploits transformer for capturing uncertain relations between audio signals and facial animations.
Specifically, the mapping network models the probability distribution of variational facial animations conditioned upon the driving audio signals, which can autoregressively transfer the audio into diverse yet realistic facial animation sequences for guiding talking face generation.
In addition, we introduce a temporally-biased mask into the mapping network, which allows to model the temporal dependency of facial animations and produce temporally coherent facial animation sequence effectively.
With the generated facial animation sequence and a source image, a generation network can directly synthesize photo-realistic talking faces. More details to be described in the ensuing subsections.

\subsection{Mapping Network}
\label{mapping_network}

\subsubsection{Facial Animations Representation}
\label{discrete_representation}

The goal of the mapping network is to convert the driving audio signals into reasonable and coherent visual signals for guiding talking face generation.
Before learning the audio-to-visual mapping, it is crucial to define a compact yet informative representation for the facial animations as visual signal. In this paper, we exploit 17 continuous Action Units (AUs)~\cite{ekman2002facial} for jointly modelling accurate mouth movements and rich facial expression details. Meanwhile, we model the head pose as a rigid 6 degrees of freedom (DoF) movement (pitch, yaw, roll and 3D translation). In this way, we encode the facial animations of each talking face frame into a 23 dimensional vector.

Existing audio-driven talking face generation methods~\cite{chen2019hierarchical, zhou2020makelttalk, wen2020photorealistic} represent facial animations with continuous value (e.g., the coordinates of 2D landmarks or the 3D face coefficients) and learn the audio-to-visual mapping in a deterministic manner. This paradigm ignores the uncertainty between audio signals and facial expressions and often leads to regress-to-mean problem~\cite{wang2021audio2head} where the synthesized faces have little variation in facial expressions.
Differently, we represent facial animations as discrete categories and formulate the output of the mapping network as a probability tensor of the categorical distribution, which allows to sample diverse yet realistic facial animations at inference.
Specifically, we uniformly discretize facial animations into $D$ constant intervals and obtain 23 $D$-dimensional one-hot vectors (including 17 AUs and 6 DoF of head movement). 
Given a facial animation vector $Y_t=[y_1, y_2, ..., y_{23}]$ of frame $t$, its discrete representation $\overline{Y}_t=[\overline{y}_1, \overline{y}_2, ..., \overline{y}_{23}]$ can be obtained as follows:
\begin{equation}
    \begin{aligned}
        \overline{y}_i = \underset{d \in D}{\mathrm{argmin}} ||y_i - y_d||_1,
    \end{aligned}
\end{equation}
where $y_i \in Y_t$, $\overline{y}_i \in \overline{Y}_t$ and $y_d$ is the centroid value of $d^{th}$ category. In our experiments, we empirically set $D$ at 500.

\begin{figure}[!t]
\begin{center}
\includegraphics[width=1.\linewidth]{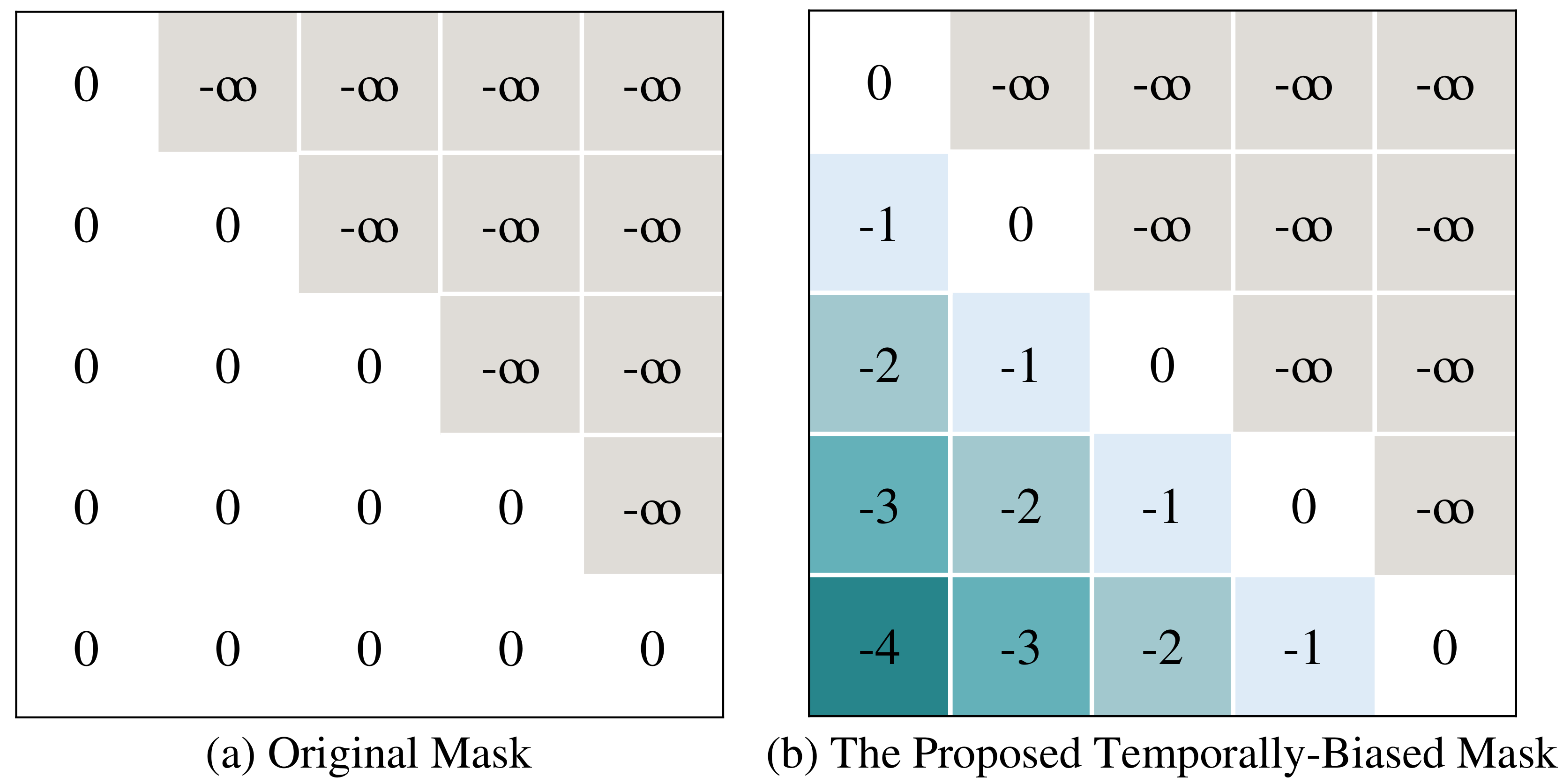}
\end{center}
\caption{
    Comparison of the original mask in~\cite{vaswani2017attention} and the proposed temporally-biased mask: Our temporally-biased mask negatively biases attention weights with a linearly decreasing penalty that is proportional to the distance between the current prediction and previously predicted animations. Our mask allows the mapping network to better model the temporal dependency of facial animations and produce temporally coherent facial animation sequence effectively. The attention weights in upper triangular are set at $-\infty$ to ensure currently predicting facial animations attend to its previously generated frames only~\cite{vaswani2017attention}.
}
\label{fig:mask_comparison}
\end{figure}

\subsubsection{Audio-to-visual Mapping}
\label{audio_visual_mapping}

Inspired by the power of transformer in modelling long-range contextual information, we introduce transformer~\cite{vaswani2017attention} into the mapping network for modelling the uncertain relations between audio and visual signals and converting the driving audio signals into facial animations in an autoregressive manner.

Autoregressive generation has been widely explored for generating sequential data, where the probability of each new observation conditions on its previous observations and the joint distribution of the whole sequence is modelled by the product of conditional distributions.
With the conditional nature of audio-driven talking face generation, we train the mapping network to autoregressively predict facial animations conditioned on both audio context and previously generated facial animations. Following Bayes' Rule, the probability of the discrete facial animation sequence $\overline{Y}_{1:T}$ under the driving audio $A_{1:T}$ can be decomposed by:
\begin{equation}
\small
    \begin{aligned}
        p(\overline{Y} | A) = p(\overline{Y}_{1:\tau} | A_{1:\tau}) \cdot \prod^{T}_{t=\tau+1}  p(\overline{Y}_t | \overline{Y}_{t-\tau:t-1}, A_{t-\tau:t}), 
        \label{formula:autoregressive} 
    \end{aligned}
\end{equation}
where we assume that facial animations at next time step $t$ depends on at most $\tau$ previous facial animations (besides the audio signals). 
Following~\cite{vaswani2017attention}, a mask is applied to the transformer to ensure the currently predicting facial animations attend to its previously generated frames only.

Once the mapping network is trained, it allows to generate diverse and plausible facial animations from the same audio in an autoregressive manner. 
Specifically, given the audio signals and previously generated animations, the mapping network first predicts the likelihood of possible facial animations for current time step.
The top-k sampling is then adopted to sample facial animations from the likelihood as the newly generated facial animations.
This process repeats until sequence of desired length is produced. 
Finally, the generated discrete facial animations $\tilde{Y}_{t}$ are converted back to continuous facial animations $\hat{Y}_{t}$ by replacing the discrete categorical label with the corresponding centroid value:
\begin{equation}
    \begin{aligned}
        \tilde{y}_i = k \mbox{ such that } \hat{y}_i = y_k,  
    \end{aligned}
\end{equation}
where $\hat{y}_i \in \hat{Y}_t$, $\tilde{y}_i \in \tilde{Y}_t$, $\hat{y}_i$ is the converted continuous value of $\tilde{y}_i$ and $y_k$ is the centroid value of the $k^{th}$ category, respectively.

\subsubsection{Improving Temporal Coherence}

As facial animations change smoothly when we talk, modelling temporal dependency of facial animations is one key factor in generating realistic talking faces. 
Without considering the temporal dependency, the vanilla transformer~\cite{vaswani2017attention} assigns equal attention weights to all previously generated facial animations while generating the new one, which often leads to abrupt facial animations in the generated faces.

Inspired by the intuition that the closer facial animation frames should have stronger correlations with the current frame and thus are more likely to affect its prediction, we introduce a temporally-biased mask into the mapping network, which biases the attention by assigning higher attention weights to the closer frames. Specifically, our temporally-biased mask negatively biases the attention weights with a linearly decreasing penalty~\cite{press2021train} that is proportional to the distance between the previously predicted frames and current prediction as illustrated in Fig.~\ref{fig:mask_comparison}. 
After that, a softmax function is employed to obtain the final attention weights as in Eq.~(\ref{transformer_basic}).
With the proposed temporally-biased mask, the mapping network is encouraged to take into account the temporal dependency of facial animations across consecutive frames while generating facial animations at current time step, which enables it to produce temporally coherent facial animation sequence effectively.
Fig.~\ref{fig:mask_comparison} shows the comparison between the original attention mask in~\cite{vaswani2017attention} and our temporally-biased mask.

\subsubsection{Loss Functions}

Given the predicted facial animation probability $p(\overline{Y} | A)$ and ground-truth discrete facial animation sequence $\overline{Y}_{1:T}$, we train the mapping network by minimizing the cross-entropy loss:
\begin{equation}
    \begin{aligned}
        L_{CE} = \frac{1}{T} \mathrm{CrossEntropy}(p(\overline{Y} | A), \overline{Y}_{1:T}).
        \label{formula:CE_loss_autoregressive}
    \end{aligned}
\end{equation}

\subsection{Generation Network}

Given a source image and the facial animations generated by the mapping network, a generation network is designed to synthesize plausible talking faces that comply with the facial animations while preserving the source identity attributes.

We design the generation network based on face-vid2vid~\cite{wang2021one} and follow existing talking face generation methods~\cite{chen2019hierarchical, zhou2021pose} to train it on video datasets~\cite{chung2018voxceleb2, chung2016lip}.
Specifically, we first randomly extract two frames from the same video track and pick one of them as the source image $I_{src}$. We then treat the other frame as the target image $I_{tgt}$ and extract the facial animation attributes from the it as the driving attributes $Z_{tgt}$. Finally, $I_{src}$ and $Z_{tgt}$ are fed into the generation network $G$, which is trained to transform $I_{src}$ to $I_{tgt}$ conditioned on $Z_{tgt}$:
\begin{equation}
    \begin{aligned}
        \hat{I}_{tgt} = G(I_{src}, Z_{tgt}),
    \end{aligned}
\end{equation}
where $\hat{I}_{tgt}$ the generated image.

With a source facial image and the facial animation sequence generated by the mapping network, the trained generation network can generate talking faces with realistic facial animations as illustrated in the right part of Fig.~\ref{fig:pipeline}.

\subsubsection{Loss Functions}
The loss for training the generation network consists of three terms:
1) the adversarial loss for improving the photo-realism of the output;
2) the reconstruction loss for penalizing the reconstruction error;
3) the attribute loss that examines whether the generated image contains desired facial animations.
The overall objective function is:
\begin{equation}
    \begin{aligned}
         \mathcal{L} = \mathcal{L}_{adv}
                      + \lambda_{rec} \mathcal{L}_{rec}
                      + \lambda_{attr} \mathcal{L}_{attr}. 
        \label{formula:generate_loss}
    \end{aligned}
\end{equation}

\noindent \textbf{Adversarial Loss: }
We adopt an adversarial loss for improving the photo-realism of the synthesized talking face images. Specifically, LSGAN~\cite{mao2017least} is employed to optimize network parameters. The adversarial loss is formulated as:  
\begin{equation}
    \begin{aligned}
        \mathcal{L}_{adv}  = 
                    \frac{1}{2} {\mathbb{E}} ({D_I}(I_{tgt})^2)
                    + \frac{1}{2} {\mathbb{E}} ((1- {D_I}( \hat{I}_{tgt}))^2),  
        \label{adv_loss}
    \end{aligned}
\end{equation}
where $D_I$ is the discriminator.

\noindent \textbf{Reconstruction Loss: }
We employ a reconstruction loss to reduce the error between the output of the generation network and the ground-truth image:
\begin{equation}
    \begin{aligned}
        \mathcal{L}_{rec} = 
                  {\mathbb{E}} ( || \phi(I_{tgt}) - \phi(\hat{I}_{tgt}) ||_1 ), 
        \label{rec_loss}
    \end{aligned}
\end{equation}
where $\phi$ is the pre-trained VGG network~\cite{simonyan2014very}.

\noindent \textbf{Attribute Loss: }
We adopt an attribute loss to encourage the generation network to generate a face image with similar facial animations as the ground-truth. Specifically, we include an auxiliary head on top of the discriminator ($D_{Z}$) to predict attributes, and apply L2 loss on the predicted attributes of both ground-truth and generated images: 
\begin{equation}
    \begin{aligned}
        \mathcal{L}_{attr} = 
                  {\mathbb{E}} ( \| D_{Z}( I_{tgt} ) - Z_{tgt} \|_2^2)
                  + {\mathbb{E}} ( \| D_{Z}( \hat{I}_{tgt} ) - Z_{tgt} \|_2^2).
        \label{attr_loss}
    \end{aligned}
\end{equation}

\section{Experiments}

\subsection{Experimental Settings}
\label{sec_settings}

\noindent \textbf{Datasets: }
We conduct experiments over two in-the-wild audio-visual datasets Voxceleb2~\cite{chung2018voxceleb2} and LRW~\cite{chung2016lip}, which are widely used in existing audio-driven talking face generation studies. 
Specifically, Voxceleb2~\cite{chung2018voxceleb2} contains over 1 million utterances for 6,112 celebrities, extracted from videos uploaded to YouTube. 
LRW~\cite{chung2016lip} contains 1,000 utterances of 500 different words, spoken by hundreds of different speakers. 
We split each dataset into training and testing sets by following the official settings.

\noindent \textbf{Implementation Details: }
The audios are pre-processed to 16kHz, then converted to mel-spectrograms with FFT windows size 1280, hop length 160 and 20 Mel filter-banks.
Then we match each video frame with audio features by sampling the mel-spectrograms with the target frame time-step in the middle.
The facial animations (i.e., the continuous AUs and head pose parameters) of each frame are extracted by OpenFace~\cite{baltrusaitis2018openface}.
$\tau$ is set at 50. 
All experiments are conducted in PyTorch~\cite{paszke2017automatic} environment with four 11GB GeForce RTX 2080 Ti GPUs.
The mapping network and the generation network are trained separately.
Please refer to the supplementary material for network training details.

\begin{figure*}[!t]
\begin{center}
\includegraphics[width=1.\linewidth]{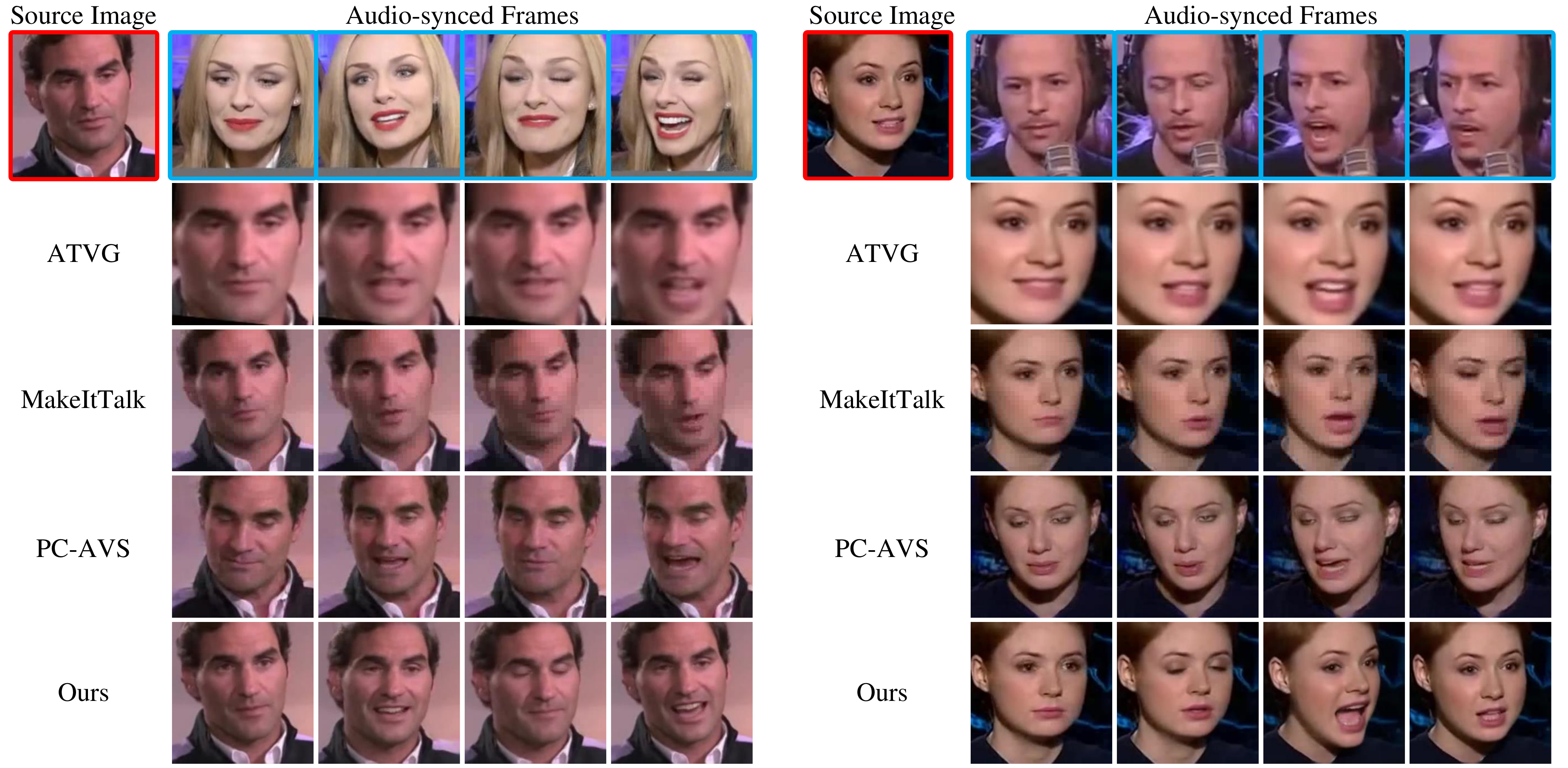}
\end{center}
\caption{
        Qualitative comparisons of DIRFA with state-of-the-art audio-driven talking face generation approaches: The first row shows the source image and audio-synced frames that provide ground-truth lip shapes. The rest rows show the synthesized faces. DIRFA can generate \textit{accurate} lip shapes with respect to the audio-synced frames, \textit{vivid} facial expressions and \textit{natural} head poses conditioned on audio only. Note PC-AVS~\cite{zhou2021pose} takes the audio-synced frames as additional input for head pose control. 
}
\label{fig:qualitative}
\end{figure*}

\begin{figure}[!t]
\begin{center}
\includegraphics[width=1.\linewidth]{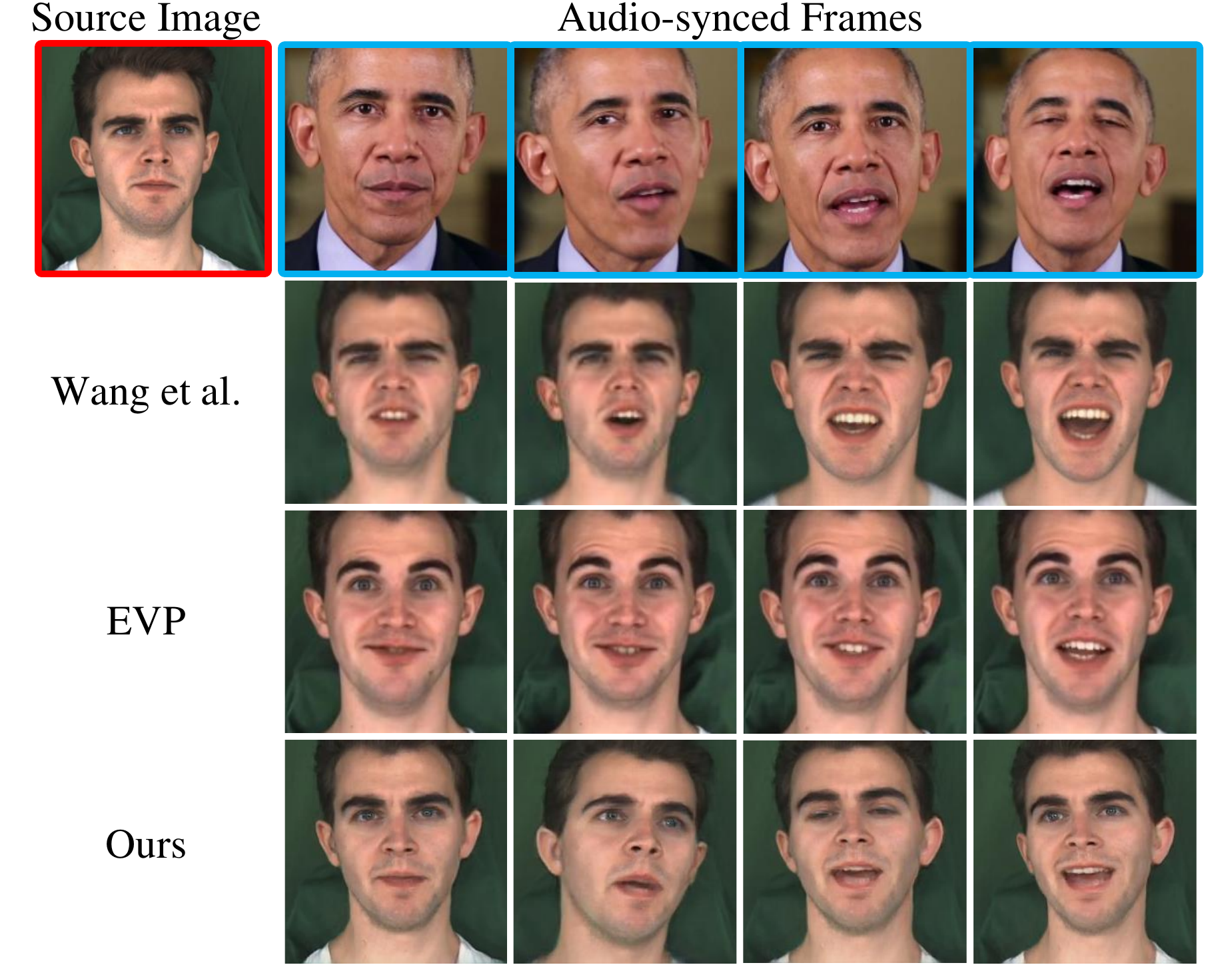}
\end{center}
\caption{
    Qualitative comparisons of DIRFA with Wang et al.'s method~\cite{wang2020mead} and EVP~\cite{ji2021audio} on MEAD~\cite{wang2020mead}: The audio-synced frames provide ground-truth lip shapes. The \textit{subject-specific training} in Wang et al.'s method~\cite{wang2020mead} and EVP~\cite{ji2021audio} helps to generate sharper facial texture, while DIRFA can generate better lip-sync and more consistent facial expressions with respect to the audio-synced frames as well as natural head movements. 
}
\label{fig:compare_mead}
\end{figure}

\subsection{Qualitative Evaluation}

We first qualitatively compare DIRFA with three state-of-the-art subject-independent audio-driven talking face generation methods, including ATVG~\cite{chen2019hierarchical}, MakeItTalk~\cite{zhou2020makelttalk} and PC-AVS~\cite{zhou2021pose}. The results are shown in Fig.~\ref{fig:qualitative}.
Note all compared methods take the source image and the driving audio as inputs (unseen in training) for generating talking faces, while PC-AVS~\cite{zhou2021pose} takes the audio-synced videos as additional input for controlling head poses.

ATVG~\cite{chen2019hierarchical} generates talking faces conditioned on 2D landmarks predicted from the audio. It can generate faces with good lip-sync with respect to the audio, but its generated head poses are static. 
Leveraging speaker-aware 3D landmarks, MakeItTalk~\cite{zhou2020makelttalk} can synthesize photo-realistic talking faces, but their head poses show slight movements only and the lip shapes are not well-aligned with the audio-synced frames. 
PC-AVS~\cite{zhou2021pose} can generate more diverse head motions and accurate lip movements than MakeItTalk~\cite{zhou2020makelttalk}, but it requires additional videos for pose guidance and struggles to preserve the identity of source image. 
Further, all three methods model audio-to-visual mapping in a deterministic manner without considering the uncertainty between audio and facial animations. This leads to regression-to-mean problem~\cite{wang2021audio2head} where the synthesized faces have little variation in facial expressions.
As a comparison, DIRFA generates talking faces with accurate lip shapes, vivid facial expressions and natural head poses conditioned on audio only.
Its superior synthesis is largely attributed to our probabilistic mapping network that can model the variational facial animation distribution conditioned upon the input audio.

We further compare DIRFA with two state-of-the-art subject-specific audio-driven emotional talking face generation methods, including Wang et al.~\cite{wang2020mead} and EVP~\cite{ji2021audio}, both of which are trained on the MEAD dataset~\cite{wang2020mead}. 
It is worth mentioning that DIRFA is trained on Voxceleb2~\cite{chung2018voxceleb2} only and directly applied to MEAD images without finetuning.
As Fig.~\ref{fig:compare_mead} shows, the subject-specific training in Wang et al.'s method~\cite{wang2020mead} and EVP~\cite{ji2021audio} helps to generate talking faces with sharper texture, but their lip shapes and facial expressions are not consistent with the audio-synced frames and the generated head poses are static.
In contrast, DIRFA generates talking faces with better lip-sync and more consistent facial expressions as well as natural head motions.
In addition, DIRFA is more flexible and generalizable, which can be applied to different persons as shown in Fig.~\ref{fig:qualitative}.

\renewcommand\arraystretch{1.2}
\begin{table*}[!t]
\small 
\renewcommand\tabcolsep{3pt}
\caption{
Quantitative comparisons of DIRFA with state-of-the-art audio-driven talking face generation approaches on datasets Voxceleb2~\cite{chung2018voxceleb2} and LRW~\cite{chung2016lip}. N/A in EBR means eye blinks are completely static in the synthesized talking faces. 
}
\centering
\begin{tabular}{|m{2.0cm}<{\centering}|
                m{2.2cm}<{\centering}|
                m{1.15cm}<{\centering}
                m{1.0cm}<{\centering}
                m{1.05cm}<{\centering}
                m{1.05cm}<{\centering}
                m{1cm}<{\centering}
                m{0.9cm}<{\centering}|
                } \hline
Dataset & Methods & PSNR $\uparrow$ & ACD $\downarrow$ & LMD $\downarrow$ & S$_{conf}$ $\uparrow$ & S$_{off}$ $\downarrow$ & EBR
\\\hline

\multirow{4}{*}{Voxceleb2~\cite{chung2018voxceleb2}} 
& ATVG~\cite{chen2019hierarchical}  & 28.79 & 0.376 & \textbf{4.28} & 5.45 & 0.57  & N/A\\
& MakeItTalk~\cite{zhou2020makelttalk} & 29.15 & \textbf{0.320} & 6.13 & 5.11 & 1.08 & 0.39\\
& PC-AVS~\cite{zhou2021pose} & 29.42 & 0.354 & 4.67 & 5.39 & 0.36 & N/A\\
& Ours & \textbf{29.56} & 0.331 & 4.45 & \textbf{5.82} & \textbf{0.25} & 0.31
\\\hline

\multirow{4}{*}{LRW~\cite{chung2016lip}} 
& ATVG~\cite{chen2019hierarchical} & 29.64 & 0.303 & 3.71 & 4.47 & 0.66 & N/A\\
& MakeItTalk~\cite{zhou2020makelttalk} & 29.93 & \textbf{0.274} & 6.62 & 3.56 & 0.82 & 0.27\\
& PC-AVS~\cite{zhou2021pose} & 30.65 & 0.288 & 3.35 & 6.21 & \textbf{0.33} & N/A\\
& Ours & \textbf{30.97} & 0.280 & \textbf{3.16} & \textbf{6.39} & 0.45 & 0.24
\\\hline

\end{tabular}
\label{tab_1}
\end{table*}

\renewcommand\arraystretch{1.2}
\begin{table}[!t]
\small 
\renewcommand\tabcolsep{3pt}
\centering
\caption{
Quantitative comparisons of DIRFA with Wang et al.'s method~\cite{wang2020mead} and EVP~\cite{ji2021audio} on MEAD~\cite{wang2020mead} dataset. N/A in EBR means eye blinks are completely static in the synthesized talking faces.
}
\begin{tabular}{|m{2.5cm}<{\centering}|
                m{1.15cm}<{\centering}
                m{1.0cm}<{\centering}
                m{1.05cm}<{\centering}
                m{1.05cm}<{\centering}
                m{1cm}<{\centering}
                m{0.9cm}<{\centering}|
                } \hline
Methods & PSNR $\uparrow$ & ACD $\downarrow$ & LMD $\downarrow$ & S$_{conf}$ $\uparrow$ & S$_{off}$ $\downarrow$ & EBR
\\\hline
Wang et al.~\cite{wang2020mead}  & 28.67 & 0.366 & 11.94 &2.38 & 1.70 & N/A \\
EVP~\cite{ji2021audio} & \textbf{28.90} & \textbf{0.307} & 8.23 & 4.21 & 0.88 & N/A \\
\hline
Ours 
& 28.42 & 0.391 & \textbf{6.45} & \textbf{5.09} & \textbf{0.63} & 0.35 
  \\\hline
\end{tabular}
\label{table:MEAD_quantitative_results}
\end{table}

\subsection{Quantitative Evaluation}

\textbf{Evaluation Metrics: }  
We perform quantitative evaluations with several metrics that have been widely adopted in existing audio-driven talking face generation studies. 
Specifically, we use peak signal-to-noise ratio (\textbf{PSNR}) to evaluate the generation quality and average content distance (\textbf{ACD})~\cite{tulyakov2018mocogan} to measure the facial identify preservation quality. 
We also adopt the landmark distance (\textbf{LMD}) around mouths~\cite{chen2019hierarchical}, confidence score (\textbf{S}$_{{conf}}$) and synchronization offset (\textbf{S}$_{{off}}$) as described in~\cite{chung2016lip} to measure the accuracy of mouth shapes and lip sync. 
Note the \textbf{S}$_{{off}}$ here refers to the absolute difference between video-audio lags of the generated and audio-synced videos. 
In addition, we measure the eye blinking rate (\textbf{EBR})~\cite{zhang2021facial} of the generated faces which is more realistic when it is similar to the average human eye blinking rate around 0.28-0.45 blinks per second~\cite{bentivoglio1997analysis}.

\renewcommand\arraystretch{1.2}
\begin{table*}[!t]
\small 
\renewcommand\tabcolsep{3pt}
\caption{
User study in mean opinion scores: Users rate videos from 1 to 5. Larger scores mean better performance. Subject-inde. and subject-spec. mean subject-independent and subject-specific, respectively.
}
\centering
\begin{tabular}{|m{2.0cm}<{\centering}|
                m{2.5cm}<{\centering}|
                m{2.0cm}<{\centering}
                m{2.2cm}<{\centering}
                m{2.2cm}<{\centering}|
                } \hline
{} & Methods & Video Realness  & Lip Sync Quality  & Expression Naturalness \\
\hline

\multirow{5}{*}{\shortstack{Subject-inde. \\ Methods}} 
& \color{gray}{Ground Truth} & \color{gray}{4.83} & \color{gray}{4.75} & \color{gray}{4.64}\\
& ATVG~\cite{chen2019hierarchical}  &1.49 &2.68 &1.35\\
& MakeItTalk~\cite{zhou2020makelttalk} &2.07 &2.11 &1.84\\
& PC-AVS~\cite{zhou2021pose} &3.16 &3.53 &2.80\\
& Ours &\textbf{3.95} &\textbf{4.02} &\textbf{3.76}\\
\hline

\multirow{4}{*}{\shortstack{Subject-spec. \\ Methods}}
& \color{gray}{Ground Truth} & \color{gray}{4.89} & \color{gray}{4.92} & \color{gray}{4.78}\\
& Wang et al.~\cite{wang2020mead} &2.57 &1.33 &1.16\\
& EVP~\cite{ji2021audio} &3.25 &3.64 &3.12\\
& Ours &\textbf{3.68} &\textbf{3.91} &\textbf{3.50}
\\\hline

\end{tabular}
\label{table:user_study}
\end{table*}

\textbf{Quantitative Results: }
Table~\ref{tab_1} shows quantitative comparisons of several subjective-independent audio-driven talking face generation methods. We can see that DIRFA achieves the best PSNR on both datasets, indicating that the DIRFA synthesized faces are sharper and more realistic than those generated by other methods. In addition, DIRFA achieves the best LMD, S$_{conf}$ and S$_{off}$ in most cases, showing that DIRFA can generate accurate lip-sync videos from audio signals. Note DIRFA obtains slightly lower ACD than MakeItTalk~\cite{zhou2020makelttalk}, largely because MakeItTalk~\cite{zhou2020makelttalk} deals with a much simpler task of generating limited facial movements while DIRFA generates various face poses and facial expressions. Further, only MakeItTalk~\cite{zhou2020makelttalk} and DIRFA can generate natural eye blinking but DIRFA can generate more realistic facial animations as illustrated in Fig.~\ref{fig:qualitative}.

We also compare DIRFA with subjective-specific audio-driven talking face generation methods. As Table~\ref{table:MEAD_quantitative_results} shows, \cite{wang2020mead} and \cite{ji2021audio} can achieve better PSNR and ACD as they conduct subject-specific training. However, their generated lip shapes are not well aligned with the audio which leads to low LMD, S$_{conf}$ and S$_{off}$. In addition, their generated eye blinks are static, which impairs the realism of the synthesized talking faces. As a comparison, DIRFA can synthesize more realistic facial animations with better lip-sync.

\begin{figure}[!t]
\begin{center}
\includegraphics[width=1.\linewidth]{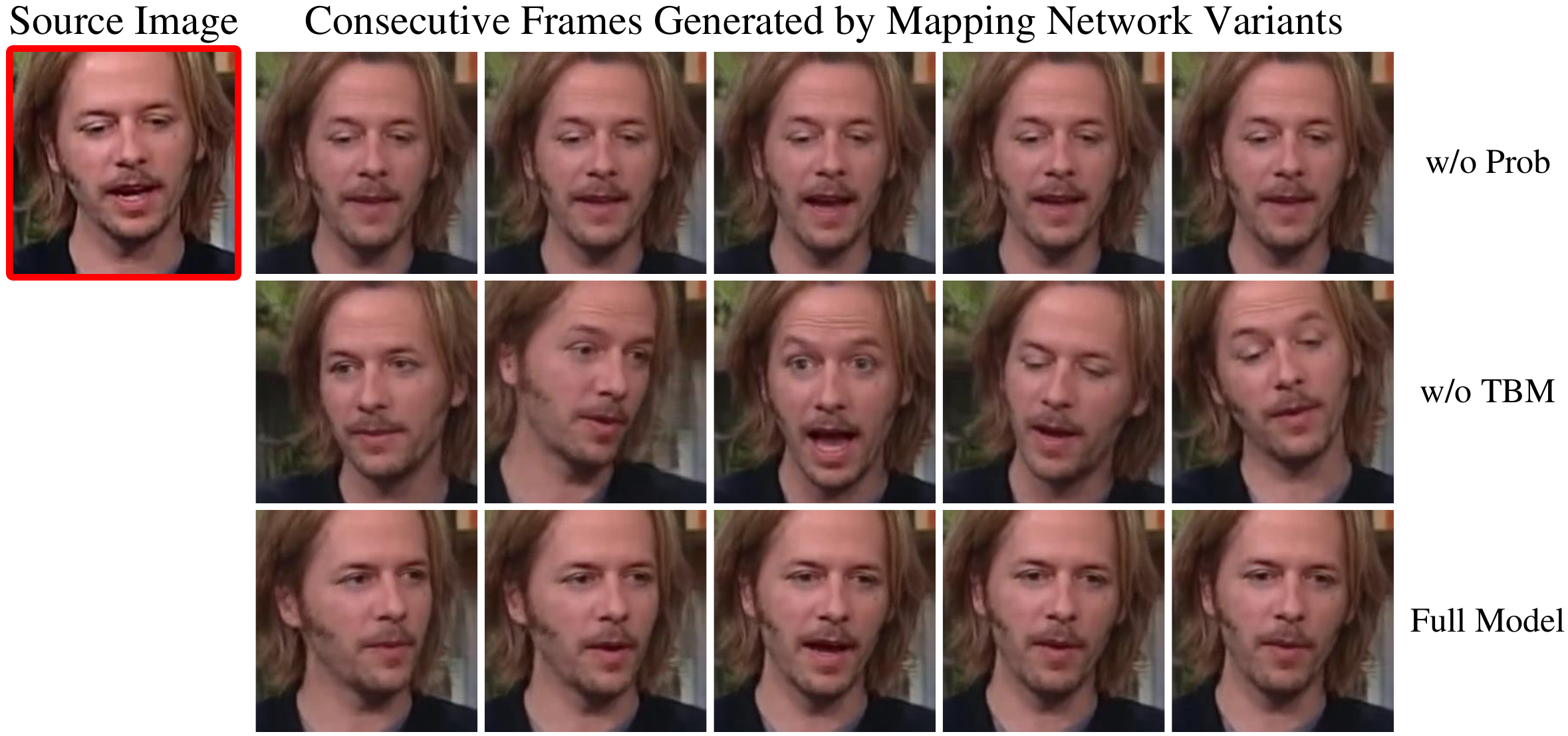}
\end{center}
\caption{
    Qualitative results of ablation study: The probabilistic design (Prob) and the temporally-biased mask (TBM) help to model audio-to-visual uncertainty and generate temporally coherent facial animations, respectively.
}
\label{fig:ablation_study}
\end{figure}

\renewcommand\arraystretch{1.2}
\begin{table}[!t]
\small 
\renewcommand\tabcolsep{3pt}
\centering
\caption{
Quantitative results of ablation study: Prob denotes the probabilistic design; TBM denotes the temporally-biased mask; N/A in EBR means eye blinks are static in the faces. 
}
\begin{tabular}{|m{2.5cm}<{\centering}|
                m{1.15cm}<{\centering}
                m{1.0cm}<{\centering}
                m{1.05cm}<{\centering}
                m{1.05cm}<{\centering}
                m{1cm}<{\centering}
                m{0.9cm}<{\centering}|
                } \hline
Methods & PSNR $\uparrow$ & ACD $\downarrow$ & LMD $\downarrow$  & S$_{conf}$ $\uparrow$ & S$_{off}$ $\downarrow$ & EBR
\\\hline
w/o Prob  & 29.52 & \textbf{0.318} & 4.71 & 5.60 & 0.43 & N/A \\
w/o TBM & 29.38 & 0.347 & 6.22 & 4.39 & 0.74 & 0.59 \\
\hline
Full Model
& \textbf{29.56} & 0.331 & \textbf{4.45} & \textbf{5.82} & \textbf{0.25} & 0.31
  \\\hline
\end{tabular}
\label{table:ablation_evaluation}
\end{table}

\renewcommand\arraystretch{1.2}
\begin{table}[!t]
\small 
\renewcommand\tabcolsep{3pt}
\centering
\caption{
Effect of different number of discrete categories $D$ on the accuracy of generated lip shapes on Voxceleb2~\cite{chung2018voxceleb2}.
}
\begin{tabular}{|m{1.5cm}<{\centering}|
                m{1.2cm}<{\centering}
                m{1.2cm}<{\centering}
                m{1.2cm}<{\centering}
                m{1.2cm}<{\centering}
                m{1.2cm}<{\centering}|} \hline
$D$ & 10 & 100 & 250  & 500 & 750 
\\\hline
LMD $\downarrow$
& 9.31  & 5.92  & 4.77 & \textbf{4.45}  & 4.68
  \\\hline
\end{tabular}
\label{table:analysis_of_D}
\end{table}

\subsection{User Study}

We conduct user studies to evaluate the perceptual realism of the generated talking faces. The subjects are presented by randomly-ordered videos synthesized by DIRFA and the compared methods (each method generates 3 videos). They are tasked to rate video quality (from 1 to 5) based on three criteria: 1) the realness of the video; 2) the lip sync quality; and 3) the naturalness of facial expressions. 
Table~\ref{table:user_study} shows the mean opinion scores of 83 users. 
DIRFA consistently outperforms competing methods under all metrics, demonstrating the effectiveness of our method.

\subsection{Discussion}

\noindent \textbf{Ablation Studies: } 
We conduct ablation studies to analyse the effectiveness of the proposed mapping network in realistic talking face generation with three variants. 
The first replaces the discrete category facial animation representations with continuous representations (`w/o Prob'). 
The second replaces the proposed temporally-biased mask with the original mask in~\cite{vaswani2017attention} (`w/o TBM').
The third is our full model (`Full Model').
The facial animation sequences produced by different mapping networks are fed to the same generation network to generate talking faces.

Fig.~\ref{fig:ablation_study} shows the qualitative results, where each row shows five consecutive generated frames.
Without the probabilistic design, the mapping network tends to learn a deterministic mapping between audio and facial animations, which suffers from regression-to-mean problem and generates static and expressionless faces. 
Without the temporally-biased mask, the mapping network produces facial animation sequences without considering the temporal dependency of facial animations, leading to generating faces with abrupt head poses. 
Including the probabilistic design and temporally-biased mask, our full model generates temporally smooth talking faces with natural facial animation transition.

Table~\ref{table:ablation_evaluation} shows the quantitative results. 
The mapping network without probabilistic design obtains better ACD as it generates faces with little variation in head poses and expressions, while our full model generates faces with realistic facial animations and achieves the best under other metrics.

\begin{figure}[!t]
\begin{center}
\includegraphics[width=1.\linewidth]{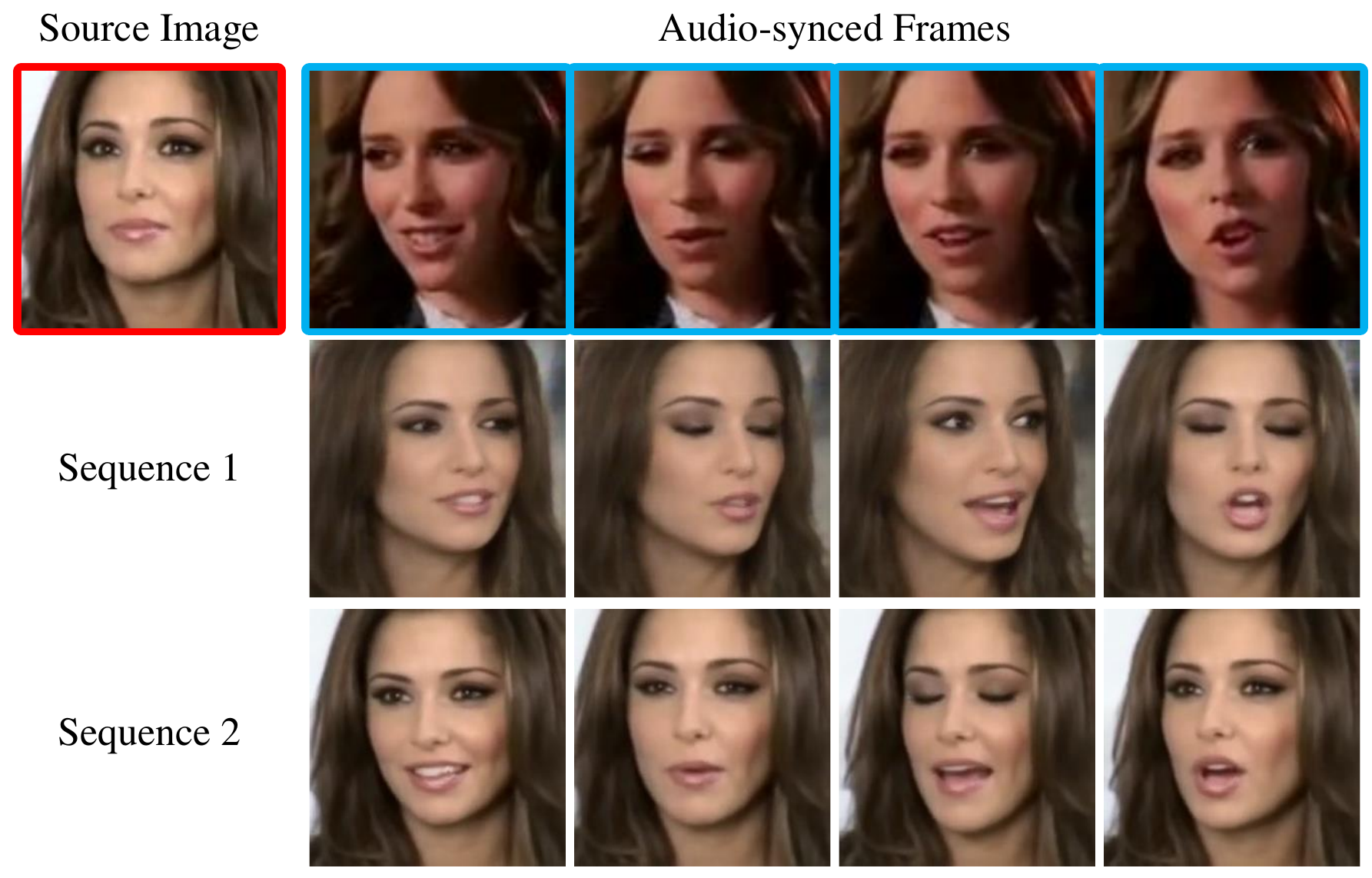}
\end{center}
\caption{
    Diverse talking faces generation: By randomly sampling facial animation sequences from the same audio, DIRFA can synthesize talking faces with \textit{accurate} and \textit{consistent} lip shapes with respect to the audio-synced frames as well as \textit{diverse} and \textit{realistic} expressions and head poses as illustrated in Rows 2 and 3.
}
\label{fig:diverse_results}
\end{figure}

\smallskip
\noindent \textbf{Discrete Category Analysis: } 
To allow sampling diverse yet realistic facial animations at inference, we represent facial animations as discrete categories and formulate the output of the mapping network as a probability tensor of categorical distribution. Specifically, we uniformly discretize facial animations into $D$ constant intervals. We examine how the change of parameter $D$ affects the accuracy of lip shapes in the generated talking faces. In the experiments, we adopt the LMD metric (lower LMD indicates better lip shape accuracy) and examine $D$ over the Voxceleb2 benchmark~\cite{chung2018voxceleb2}.

As Table~\ref{table:analysis_of_D} shows, the accuracy of lip shapes is severely degraded when $D$ is very small (i.e., 10). We conjecture that a small $D$ significantly restricts possible states of facial animations, which leads to inaccurate and unnatural lip movements in the generated faces. When we increase $D$, the LMD first drops and then converges to a plateau. The best LMD is obtained when $D$ is set at 500 (default setting in our experiments).

\smallskip
\noindent \textbf{Diverse Talking Faces Generation: } 
DIRFA can generate diverse talking faces from the same driving audio. 
Thanks to our designed probabilistic mapping network, we can easily obtain diverse facial animation sequences by randomly sampling facial animations from the predicted likelihood (as discussed in section~\ref{audio_visual_mapping}). 
The generation network can then generate diverse talking faces with the same source image plus diverse animation sequences as shown in Fig.~\ref{fig:diverse_results}.

\smallskip
\noindent \textbf{Limitations and Future Work: }
While our proposed DIRFA can generate audio-driven talking faces with diverse yet realistic facial animations, the current design generates talking faces from audio via a fully automatic pipeline. It remains an open challenge to incorporate user interaction for controlling certain desired synthesized facial animations. We will explore it in our future work.

\section{Conclusion}
This paper presents DIRFA, an audio-driven talking face generation method that can generate talking faces with diverse yet realistic facial animations from the same driving audio.
We design a transformer-based probabilistic mapping network to model the uncertain relations between audio and visual signals and introduce a temporally-biased mask into the mapping network to convert audio into temporally smooth facial animations in an autoregressive manner.
Our generation network then takes the generated facial animations and a source image as input to synthesize talking faces. 
Extensive experiments show that DIRFA can generate talking faces with accurate lip movements, vivid facial expressions and natural head poses. 

\bibliographystyle{elsarticle-num}
\bibliography{egbib}

\end{document}